\documentclass{article}
\usepackage{graphicx} 
\usepackage{authblk}

\title{CHS-SQL: A Text-to-SQL approach based on Confidence-Guided Heuristic Search Schema Linking process.}
\author[1,2,3]{Minghao Yang}
\author[1,2,3,4]{Yanjun Xu\thanks{Corresponding author: Yanjun Xu (b. 1979), Master's Supervisor, Research Senior Engineer, and Ph.D. Candidate. Research interests: digital supply chain, intelligent port-industry-city integration, artificial intelligence, and traffic information engineering and control. Email: xu.yanjun@coscoshipping.com}}
\affil[1]{Hainan COSCO Shipping Technology Co., Ltd.}
\affil[2]{Haikou Key Laboratory of Digital Supply Chain Joint Innovation}
\affil[3]{Engineering Technology Research Center for Intelligent Port-Industry-City Synergistic Development}
\affil[4]{Artificial Intelligence Application Research and Development Center}
\affil[ ]{ \{yang.minghao3, xu.yanjun\}@coscoshipping.com}
\date{March 2026}

\usepackage{booktabs}
\usepackage{tabularx}
\usepackage{multirow}
\usepackage{makecell}
\usepackage{array}
\usepackage{graphicx}
\usepackage{float}
\usepackage{adjustbox}
\usepackage{amsmath}
\usepackage{amsfonts}
\usepackage{pgfplots}
\usepackage{tikz}
\pgfplotsset{compat=1.18}

\begin{document}

\maketitle
\begin{abstract}
Recently, there have been several works in the Text-to-SQL  domain that utilize Small  Language Models (SLMs) \textsuperscript{\cite{wang2025comprehensive}} for training. These approaches achieve performance close to that of large models in generating SQL, using only the computational power of a single NVIDIA RTX 4090 GPU, while also ensuring data security. Most existing methods filter out redundant tables and columns during Schema Linking to improve Text-to-SQL accuracy. However, they do not consider the precision-recall trade-off when selecting the candidate schema subset. Our research found that both the precision and recall of Schema Linking directly affect the final SQL accuracy. Therefore, we propose a novel framework for efficiently fine-tuning SLMs on Text-to-SQL tasks,  CHS-SQL, that not only balances precision and recall but also improves overall performance on Text-to-SQL tasks. Its main innovation lies in the Schema Linking phase, where a heuristic search combined with model internal confidence is employed to achieve an optimal precision-recall trade-off.  This elaborated mechanism maximizes the precision of relevant schema candidates for the generated SQL queries while suppressing irrelevant noise. The same strategy is further applied during SQL generation to refine candidate queries while helping the SLM to avoid trapping in a local optimum. Our method achieves state-of-the-art (SOTA) results on Text-to-SQL tasks via SLMs.
\end{abstract}

\section{Introduction}

The text-to-SQL task enables data analysts to retrieve structured data from databases through natural language interaction, eliminating the need to manually comprehend database schema structures or devise SQL query statements. This significantly accelerates the development of BI reports and enhances the productivity of data analysts.

Mainstream approaches for LLM(Large Language Model)-based SQL generation can be broadly categorized into three paradigms: prompt engineering, efficient fine-tuning, and multi-agent collaboration. Currently, numerous studies have demonstrated that filtering out irrelevant schema information can effectively enhance the performance of LLM-based text-to-SQL systems. To enhance generation accuracy, several of these methods employ a Schema Linking process to filter out redundant schema information. 

Common prompt engineering techniques include Few-Shot Learning \textsuperscript{\cite{nan2023enhancing}} and CoT (Chain-of-Thought ) prompting \textsuperscript{\cite{wei2022chain}}. Few-Shot Learning improves the model's adaptability to the text-to-SQL task by providing it with several input-output examples, thereby enhancing SQL generation accuracy.  CoT techniques decompose complex queries into sequential sub-tasks, guiding the large model to generate intermediate reasoning steps and gradually solve intricate problems. Methods like DAIL-SQL\textsuperscript{\cite{gao2023text}} and DIN-SQL\textsuperscript{\cite{pourreza2023din}} rely primarily on the intrinsic capabilities of large language models (LLMs) and their comprehension of Schema Linking instructions within prompts to filter out irrelevant schema items. Although this approach leverages the LLM's reasoning ability, it often results in low recall, as the model may overlook some relevant schema components—especially those mentioned implicitly or indirectly in the natural language question.

Efficient fine-tuning methods adapt large models to specific downstream tasks by updating only a small subset of parameters. A prevalent strategy in this category adopts an ensemble-like approach, employing two specialized fine-tuned models—one for Schema Linking and another for SQL generation—to collaboratively accomplish the text-to-SQL task. For instance, RESD-SQL\textsuperscript{\cite{li2023resdsql}}, a discriminative-model-based SOTA method, first scores all tables and columns for relevance and then rigidly selects the top-4 tables and top-5 columns as input for the SQL generation model. While effective, this fixed-threshold strategy suffers from suboptimal precision, as it may include marginally relevant or even irrelevant schema elements due to its inflexible selection criteria.

Multi-agent collaboration frameworks decompose the complex text-to-SQL process into distinct sub-tasks, coordinating multiple purpose-built agents to achieve the final SQL output.  The MAC-SQL\textsuperscript{\cite{wang2025mac}} framework achieves state-of-the-art (SOTA) performance on the BIRD benchmark. It employs a carefully designed selector agent to filter out irrelevant table and column names, effectively reducing schema noise. Within the MAC-SQL selector agent, if a table is deemed relevant to the question and contains ten or fewer columns, all of its columns are retained and passed to the subsequent stage after ranking. While this strategy helps preserve potentially relevant attributes and thus improves recall, it inevitably introduces irrelevant columns from such tables, thereby degrading the overall precision of the Schema Linking step. However, it does not quantitatively measure the precision-recall trade-off in the Schema Linking process, nor does it explicitly address the balance between the two.

However, many prevalent LLM-based approaches fail to adequately quantify the precision-recall trade-off inherent in the Schema Linking process. They typically lack a quantitative assessment of the precision-recall trade-off inherent in the Schema Linking results, thereby limiting their ability to systematically drive further improvements in SQL generation performance. Therefore, effectively addressing the precision-recall trade-off in the Schema Linking phase is a critical factor in advancing the overall performance of Text-to-SQL systems. Achieving an adaptive and balanced selection of schema elements, rather than relying on fixed thresholds or uncalibrated model judgments, is essential for maximizing both accuracy and robustness.

CHS-SQL proposes an SLM-based framework for the Text-to-SQL task. In the Schema Linking phase, it introduces a novel approach that combines Beam Search with model internal confidence to filter out irrelevant table and column names. The Beam Search method effectively increases the recall of the candidate schema subset during the Schema Linking phase. In contrast, filtering irrelevant schema elements using model internal confidence effectively improves precision. By combining the hyper parameters Beam Width and Trace Confidence, we can achieve quantitative control over the Schema Linking process. This integration enables a more refined and adaptive selection process, effectively achieving an optimal precision-recall trade-off in a principled manner.

During the SQL generation phase, the CHS-SQL method utilizes Schema Linking results that balance precision and recall, as input for SQL generation to effectively improving SQL accuracy. Additionally, Beam Search is employed during SQL inference to help the SLMs avoid becoming trapped in local optima, while model internal confidence is used to select the highest-confidence SQL statement from multiple generations of SLMs. The combined effect of these multiple methods ultimately improves SLMs performance on Text-to-SQL task.

This method, CHS-SQL,  achieves SOTA performance on both the SPIDER and BIRD benchmarks through fine-tuning  SLMs on the Text-to-SQL task, thereby strongly demonstrating the effectiveness of its approach to balancing precision and recall. To facilitate the reproducibility of our proposed method, all source codes have been made publicly available on our GitHub repository.\footnote{https://github.com/ymhaolove-maker/chs-sql.git}

\section{\textbf{Related Work}}
In recent years, due to the rapid development of deep learning technology and the availability of large amounts of training data, the Text-to-SQL task has made significant progress. 

One of the early representative models is the Seq2Seq \textsuperscript{\cite{sutskever2014sequence}} model, which is a general model architecture used to solve sequence-to-sequence mapping problems. The encode-decoder structure of Seq2Seq enables the processing of input and output texts of different sequence lengths, and this architecture to some extent reflects the human information transmission model. The Seq2Seq  model has achieved SOTA results in tasks such as text translation, text summarization, and conversational question-answering. Naturally, in early research, the task of converting natural language to SQL was considered a sequence-to-sequence task, and for a long time, models based on the Encode-decoder architecture became the mainstream approach for Text-to-SQL. 

Based on the Encoder-Decoder architecture, Text-to-SQL first learns the joint representation of natural language questions and user's local database metadata through the Encoder. In this process, the Encoder, based on the user's local database metadata, identifies tables, columns, or numerical conditions relevant to the question, known as Schema Linking\textsuperscript{\cite{qin2022survey}}. The Decoder then generates the corresponding SQL query based on the learned question and metadata representation from the Encoder, followed by validation and error correction\textsuperscript{\cite{qin2022survey}}.

In the encoder part, representative works include IRNet \textsuperscript{\cite{guo2019towards}}, which utilizes NL encoder and Schema encoder to achieve question representation and Schema Linking. The encoder selects BiLSTM\textsuperscript{\cite{graves2012long}} and BERT\textsuperscript{\cite{devlin2019bert}} as base model, and it also applies an Attention mechanism. The encoder of RYANSQL\textsuperscript{\cite{choi2021ryansql}} uses a CNN model to capture local information, and a transformer model to learn the context of the question. In recent years, some works have used a graph to express the relationship between questions and metadata during Schema Linking, enhancing the encoder's learning and representation of relationships. RASAT\textsuperscript{\cite{qi2022rasat}}, based on the T5 model, generates an interaction graph from the input question sequence and Schema information. It then uses two trainable lookup tables capable of generating graph relationship embeddings to achieve relation-aware attention. Similar works include LGESQL\textsuperscript{\cite{cao2021lgesql}} and SADGA\textsuperscript{\cite{cai2021sadga}} . RESDSQL \textsuperscript{\cite{li2023resdsql}} initially uses a ranking-enhanced encoder to filter out irrelevant tables and fields, thereby alleviating the difficulty of Schema Linking in the SQL parsing process.

On the Decoder side, it is mainly divided into sketch-based method\textsuperscript{\cite{xu2017sqlnet}\cite{hwang2019comprehensive}\cite{hui2021improving}} and generation-based method. The sketch-based method breaks down the SQL generation task into different sub-modules according to syntax, then fills the slots in different sub-modules with metadata information or column values, at last integrates all sub-modules into the final SQL statement. For example, SQLNet\textsuperscript{\cite{xu2017sqlnet}} splits SQL statement generation into “WHERE” clause and “SELECT” clause as two sub tasks. It then uses a Sequence-to-Set model and column attention mechanism to predict and fill slot contents, forming the final SQL statement. However, the drawback of sketch-based methods is that templates must be predefined, leading to poor generalization. The generation-based method\textsuperscript{\cite{guo2019towards}\cite{wang2020rat}\cite{huang2021relation}} uses abstract syntax trees with syntactic rules as prior knowledge to generate SQL statements. PICARD\textsuperscript{\cite{scholak2021picard}} utilizes a Constrained Decoder to find valid output sequences by rejecting inadmissible tokens at each decoding step.

In the process of using natural language to generate SQL, there exists a mismatch problem in the literature\textsuperscript{\cite{qin2022survey}}. This is because SQL language is originally designed as a structured programming language for querying databases, making it difficult to map various expressions of human intent in natural language, unlike translation tasks. To bridge this gap between natural language and SQL, some works have designed SQL intermediate representations(IR)\textsuperscript{\cite{qin2022survey}}. IRNet \textsuperscript{\cite{guo2019towards}},Syntaxsqlnet\textsuperscript{\cite{yu2018syntaxsqlnet}} and NaturalSQL\textsuperscript{\cite{gan2021natural}}generate an IR based on the natural language query and Schema information, and then derive the corresponding SQL query based on the IR. Similar works include RESDSQL\textsuperscript{\cite{li2023resdsql}}, which first generates an SQL skeleton and then lets the decoder fill in the skeleton to generate the SQL statement.

The release of  LLMs represented by OpenAI GPT-3.5 amazed the world, showcasing the huge potential of these models in semantic understanding and text generation. More and more work has shifted towards LLMs, including tasks like Text-to-SQL. Din-SQL\textsuperscript{\cite{pourreza2023din}} and C3\textsuperscript{\cite{dong2023c3}} decompose Text-to-SQL into different subtasks and designing corresponding Prompt templates to guide the LLMs in generating correct SQL queries. These Prompting approaches eliminate the need for pre-training or fine-tuning of LLMs, saving the effort of preparing substantial training data and consuming significant computational resources. Dail-SQL\textsuperscript{\cite{gao2023text}} employs supervised fine-tuning in different LLMs to enhance their performance on Text-to-SQL tasks.

Numerous prior works have demonstrated that reducing irrelevant schema information effectively improves the accuracy of SQL generation\textsuperscript{\cite{zhang2024benchmarking}}. Consequently, many recent LLM-based Text-to-SQL approaches treat Schema Linking as a dedicated sub task. However, these methods generally fail to address the retrieval efficiency of schema elements—specifically, the precision–recall trade-off.

For instance, RESDSQL\textsuperscript{\cite{li2023resdsql}} fixes its output to the top-4 highest-scoring tables and top-5 attributes, regardless of actual relevance. DTS-SQL\textsuperscript{\cite{pourreza2024dts}} first uses a fine-tuned large language model to select relevant tables and then includes all columns from those tables in the final schema input. MAC-SQL\textsuperscript{\cite{wang2025mac}}, on the other hand, retains all tables but limits each to its top-6 most relevant attributes. 

While these strategies achieve high recall in retrieving metadata, they suffer from low precision, introducing substantial redundant or irrelevant schema elements that negatively impact downstream SQL generation.

\section{Preliminaries}

\subsection{LLM-based Text-to-SQL}

In an LLM-based Text-to-SQL system, LLMs are employed to facilitate the transformation of natural language questions into executable SQL queries. Specifically, Let Q be a natural language question and S be the database schema. S is defined by a tuple S = ( T , C, K ), where T represents multiple tables, C  represents columns, and K represents foreign key relationships. The goal is to produce a SQL statement Y which is executable and accurately represents the intent of Q. Given the prompt template P ( Q, S ), the generation process of the SQL statement Y by an SLM $M$ can be formally defined as a conditional probability distribution: 
 \begin{equation} 
 \mathbb{P}_{\mathcal{M}}(\mathcal{Y} \mid \mathcal{P}(\mathcal{Q}, \mathcal{S}))=\prod_{i=1}^{|\mathcal{Y}|} \mathbb{P}_{\mathcal{M}}\left(\mathcal{Y}_i \mid \mathcal{P}(\mathcal{Q}, \mathcal{S}), \mathcal{Y}_{1: i-1}\right) 
 \end{equation}  
Here, LLM autoregressively generates each token, $Y_i$  denotes the \textit{i}-th token of the SQL statement $Y$, and $|Y|$denotes the length of the query $Y$.

\subsection{Schema Linking}
Schema Linking is a crucial step in Text-to-SQL pipelines. Its goal is to retrieve the relevant tables and columns of a target database for a user’s query while disregarding irrelevant ones. $M_s$  is a  fine-tuned SLM for schema linking that executes the prompt $P_s$.  For the input query $Q$ and database schema $S$, it outputs $\hat{S}$  which is the schema subset necessary for generating the target SQL statement.
 \begin{equation} 
 \mathbb{P}_{\mathcal{M_s}}(\mathcal{\hat{S}} \mid \mathcal{P_s}(\mathcal{Q}, \mathcal{S}))=\prod_{i=1}^{|\mathcal{\hat{S}}|} \mathbb{P}_{\mathcal{M_s}}\left(\mathcal{\hat{S}}_i \mid \mathcal{P_s}(\mathcal{Q}, \mathcal{S}), \mathcal{\hat{S}}_{1: i-1}\right) 
 \end{equation}  

\section{Methodology}

\subsection{overview}
Previous research has confirmed that pruned Schema Linking results help improve the accuracy of SQL generation\textsuperscript{\cite{taniguchi2021investigation}} . Current Text-to-SQL methods mainly rely on the power of the LLMs itself to filter out redundant information. However, these methods do not fully consider the balance between precision and recall in the Schema Linking process. We found that finding the right balance between precision and recall actually benefits the final SQL generation task. Therefore, we developed the CHS-SQL approach.

CHS-SQL is an innovative SLM-based framework for Text-to-SQL that embodies this decomposition principle.  It explicitly splits the task into two sub tasks: Schema Linking and SQL generation. Each sub task is handled by SLMs that have undergone parameter-efficient fine-tuning, complemented by ingenious designed mechanisms to assess the confidence of intermediate outputs. 

In the Schema Linking phase, CHS-SQL employs a hybrid strategy combining Beam Search with model internal confidence to filter out irrelevant schema elements (i.e., tables and columns). The Beam Search method helps find more relevant schema subset, which increases recall rate. At the same time, using  Trace Confidence filtration helps remove unnecessary or wrong results, which improves precision rate. Both the Beam Width and the Confidence Threshold value can be adjusted as hyper parameters to balance precision and recall . By changing these two values, we can precisely control the balance between precision and recall. The resulting high-confidence schema subset is then passed as input to the SQL generation module.

In the SQL generation phase, Beam Search is first applied to enhance the precision of candidate queries. Subsequently, multiple diverse SLMs independently generate SQL statements based on the filtered schema. The final output is selected from this ensemble by choosing the query with the highest model internal confidence—effectively identifying the most stable and reliable generation trace.

Implementation details are elaborated in the following sections.

\begin{figure}
    \centering
    \includegraphics[width=1\linewidth]{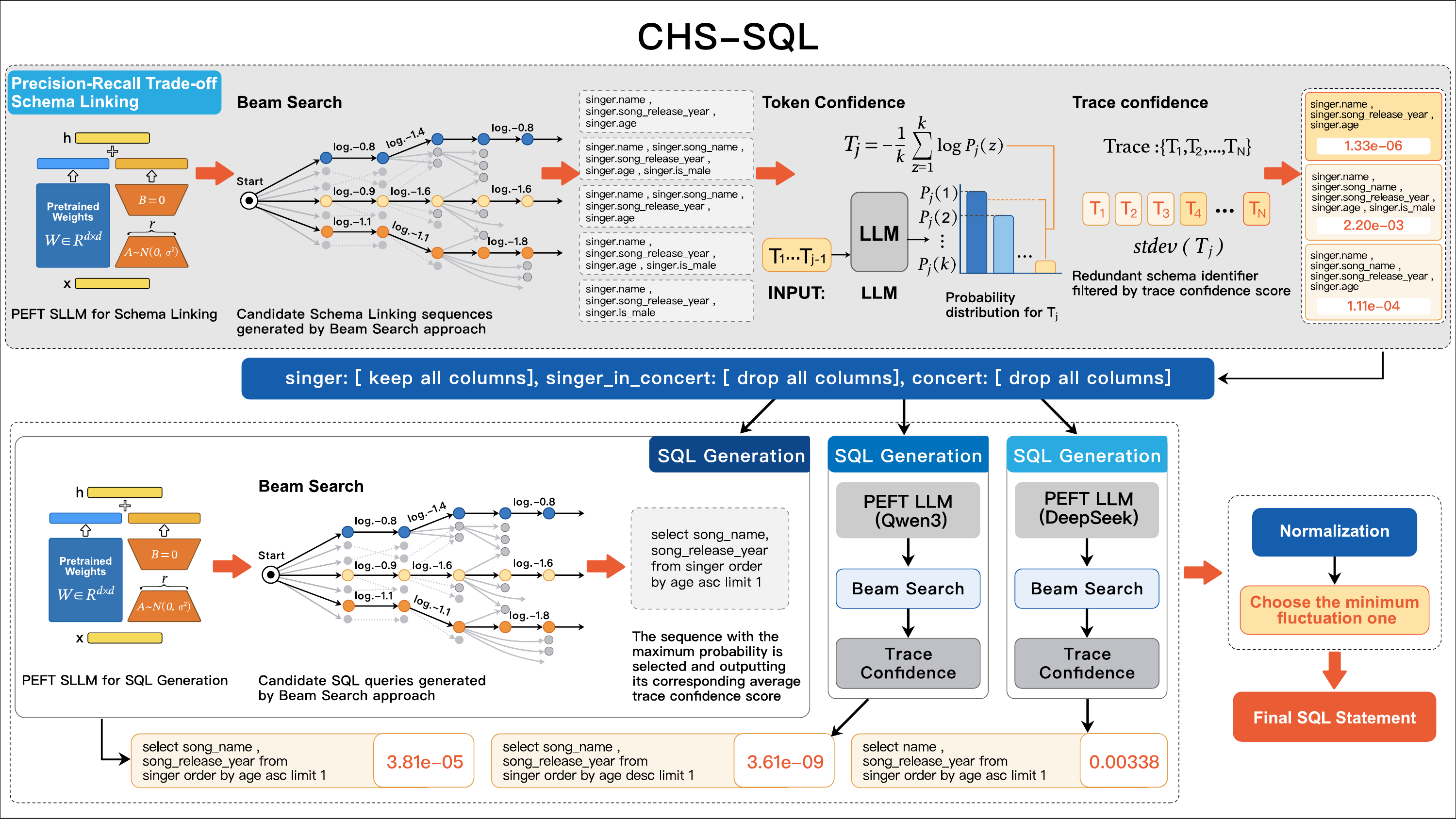}
    \caption{CHS-SQL Structure. During the Schema Linking phase, CHS-SQL employs the beam search method to increase the recall of relevant schema information, and subsequently utilizes Trace Confidence filtration to regulate precision, ultimately achieving an optimal precision-recall trade-off for the Schema Linking task. During the SQL generation phase, the Beam Search method is adopted to prevent the generated sequences from becoming trapped in local optima. Furthermore, multiple SLMs generate candidate SQL simultaneously, after which the SQL statement with the highest generation stability is selected based on the Trace Confidence score.
}
    \label{fig:placeholder}
\end{figure}

\subsection{{precision and recall trade-off Schema Linking}}
In the Schema Linking phase, we employ a SLMs that has been efficiently fine-tuned to filter out tables and columns irrelevant to the  question from the full schema. This filtering process enhances the accuracy of the subsequent SQL generation stage. In the filtering process during Schema Linking, the trade-off between precision and recall is explicitly controlled by the Beam Search approach and the Trace Confidence filtration.  This phase consists of three key steps: PEFT (Parameter-Efficient Fine-Tuning ) \textsuperscript{\cite{wan2023efficient}}, Candidate sequence generation and Candidate sequence selection. 

\subsubsection{{{PEFT}}}
We employ a PEFT approach to enhance the performance of SLMs on the Schema Linking sub task. 

Formally, for the Schema Linking task, given a database schema $D_i$ and a natural language question $q_i$ as input to the SLMs, the model outputs the set of tables $T_i$ and columns $C_i$ that are referenced in the ground-truth SQL query corresponding to $ (q_i, D_i)$. 

Our objective is to train the SLMs on a dataset  $\tau = {(q_i, D_i, T_i, C_i)}$ using an efficient fine-tuning method such that the empirical loss is minimized:

\begin{equation}
\min_{\sigma, \mathcal{M}^*} \frac{1}{|T|} \sum_{i=1}^{|T|} \mathcal{L}_{\mathcal{M}^*} \big( \sigma(q_i, D_i), T_i, C_i \big)
\end{equation}

The loss function  $\mathcal{L}$  guides parameter optimization by comparing the predicted schema elements against the ground-truth schema information. Let  $\sigma $ denote the input representation fed into the SLMs, which encodes both the natural language question and the structured schema information. We denote the resulting SLMs as  $ \mathcal{M}^* $, which supervised fine-tuning via LoRA(Low-Rank Adaptation)\textsuperscript{\cite{hu2022lora}}   on dataset   $\tau$ .

\subsubsection{{{Candidate sequence generation}}}
CHS-SQL employs Beam Search—a widely adopted decoding strategy in LLM-based text generation—to produce candidate Schema Linking sequences. Beam Search \textsuperscript{\cite{mikolov2013efficient}} is a heuristic search algorithm that maintains a fixed-size set of top-scoring partial sequences (beams) at each decoding step. In contrast to the default greedy search strategy—which selects the token with the highest probability at each step and thus risks converging to a locally optimal output—Beam Search explores a broader hypothesis space and is more likely to approximate the globally optimal sequence. 
At each decoding step, Beam Search maintains the top-k candidate sequences with the highest cumulative scores, where the score of a sequence is typically computed as the sum of log-probabilities of its constituent tokens.

Here, $K$ is referred to as the beam width, and Beam Search ultimately produces the top-k sequences with the highest cumulative scores as the output for Schema Linking:
\begin{equation}
BeamSearch=TopK(Score(w_1,w_2,...))
\end{equation}
By employing Beam Search to generate numerous reasoning traces, SLMs inherently enhance the recall of the corresponding tables and columns produced during the Schema Linking phase.

\subsubsection{{{Candidate sequence selection}}}
Compared to the default greedy decoding strategy employed by SLMs during token generation, Beam Search returns multiple reasoning traces. While this approach improves recall with respect to the ground-truth tables and columns in Schema Linking, it often comes at the cost of reduced precision due to the inclusion of spurious or low-quality candidates.

Inspired by Kang et al. (2025)  \textsuperscript{\cite{kang2026scalable}}, who demonstrate that the quality of reasoning traces in LLMs can be assessed through the model’s internal token-level probability distributions, we propose to evaluate and filter the reasoning traces generated by Beam Search. Specifically, we introduce two confidence-based metrics—Token Confidence\textsuperscript{\cite{fu2025deep}} and Trace Confidence \textsuperscript{\cite{fu2025deep}}—to quantify the reliability of each reasoning trace and discard those deemed low-quality, thereby enhancing the precision of the Schema Linking output produced by SLMs.

The token confidence for a generated token $ T_j $  at position  $j$ is defined as:
\begin{equation}
   T_j = -\frac{1}{|k|} \sum_{k=1}^{|k|} \log P_j(z_k) 
\end{equation}
where $ T_i $  denotes the negative average log-probability of the top-k tokens at position $i$. 

The average trace confidence $T_{avg} $ measures the overall distribution statistics of a reasoning trace, and is defined as follows:
\begin{equation}
T_{avg} = \frac{1}{N} \sum_{j=1}^{N} T_j
\end{equation}
where N is the total number of generated tokens. Furthermore, we employ the standard deviation of token-level probability distributions to assess whether the SLMs exhibits substantial variability during generation. This metric serves as a quantitative indicator of reasoning trace quality, known as Trace Confidence.
\begin{equation}
TC = \sqrt{\frac{1}{N} \sum_{j=1}^{N} (T_j-T_{avg})^2}
\end{equation}
Low trace confidence corresponds to greater model certainty, which in turn leads to more accurate Schema Linking predictions. Conversely, high trace confidence indicates higher predictive entropy and reflects uncertainty in the generated output.

\subsubsection{{{ Integrated Approach }}}
Overall, CHS-SQL achieves a precision–recall trade-off in the Schema Linking phase through three key components: PEFT(parameter-efficient fine-tuning), Beam Search–based generation of candidate Schema Linking sequences, and a confidence-aware selection mechanism grounded in the model’s internal token probabilities. Formally, the procedure is defined as follows:

Let $H= \{ h_1, h_2, \dots, h_k \}  $ denote the top-k Schema Linking candidate sequences generated by: 
\begin{equation}
 H =BeamSearch_k (\mathcal{M}^*,\sigma(q, D))
\end{equation}
where each hypothesis  $ h_i \in \mathcal{H} $ is a sequence of predicted table-column tokens. For the i-th candidate sequence $  h_i = (h_{i,1}, \dots, h_{i,|h_i|}) $,  its confidence score  $TC(h_i)$  is computed as: 

\begin{equation}
TC(h_i) = \sqrt{\frac{1}{|h_i|} \sum_{j=1}^{|h_i|} (T_j-T_{avg(i)})^2}
\end{equation}

We retain only those candidates whose confidence less than a  threshold $ \theta $:
\begin{equation}
 \mathcal{H^*} = \{ h_i  \mid TC(h_i) \leq \theta, h_i \in \mathcal{H} \}. 
\end{equation}
We use the Max-Gap Midpoint Sampling method to find the best value for the threshold  $ \theta $ .

The final Schema Linking output is then constructed by taking the union of all table-column elements extracted from the filtered candidates: 
\begin{equation}
 \mathcal{\hat{D}} = \bigcup_{h_i \in \mathcal{H}^*} \operatorname{Parse}(h_i)
\end{equation}
where $ \operatorname{Parse}(h_i)$ denotes the set of structured schema elements (e.g., \verb|table.column|) derived from the sequence $ h_i $. 

The beam width $k$ and confidence threshold $\theta$ are chosen to maximize table/column recall and precision:
\begin{equation}
(k^*,\theta^*)=
\arg\max_{k\in K,\ \theta\in\Theta}
\left(
R_{\text{tab}} + P_{\text{tab}} + R_{\text{col}} + P_{\text{col}}
\right).
\end{equation}

In our experiments, we observed that among the structured schema elements  selected during the candidate sequence selection stage, the table-level information is significantly more effective—achieving a better precision–recall trade-off—and leads to a notable improvement in SQL generation performance. In contrast, the column elements selected by the same procedure yield only marginal gains in downstream SQL accuracy.

\begin{equation}
\hat{D}_i = \{table_1.col_1, table_1.col_2, table_2.col_1,……,t_i.c_i\}
\end{equation}

To address this limitation, we adopt an enhanced strategy: instead of relying solely on the individually selected columns from the previous stage, we include all columns from the selected tables as input to the subsequent SQL generation phase. This design effectively boosts the recall of relevant columns, thereby improving the completeness of the schema context available for SQL synthesis.
\begin{equation}
Cols(t_i) = \{col_1, col_2,col_3,…… ,c_i\}
\end{equation}
\begin{equation}
D^*  = \{t.c |  c \in  Cols(t)\}
\end{equation}

Experimental results confirm that this approach yields significant performance gains. While this strategy resembles that of DTS-SQL—which also uses all columns from retrieved tables—our method differs critically in how candidate tables are selected. Specifically, DTS-SQL relies entirely on greedy search outputs from a fine-tuned model for table retrieval, whereas we propose a delicately designed mechanism to explicitly optimize the precision–recall trade-off for table selection. Our results demonstrate the clear superiority of this approach.

\subsection{{SQL Generation}}
Through the delicately designed precision–recall trade-off Schema Linking procedure described above, our approach maximizes the identification of question-relevant tables and columns while effectively filtering out redundant schema elements. This mitigates the adverse impact of extraneous information on SQL generation. The resulting structured schema elements are then incorporated as prompt input to the supervised fine-tuning SLM responsible for SQL generation.

During SQL generation, we ensemble multiple diverse SLMs to produce a rich set of candidate outputs. Inspired by the Schema Linking stage, we employ Beam Search to generate multiple SQL queries and apply a reasoning trace confidence based filtering mechanism—derived from the model’s internal token-level probabilities—to select the highest-quality final SQL query. 

Analogous to the Schema Linking pipeline, the SQL generation process consists of three stages: (1) PEFT , (2) Candidate SQL Generation, and (3) Candidate SQL Selection. Drawing on ensemble learning principles, we fine-tuning several heterogeneous SLMs  to capture complementary linguistic and structural patterns in the PEFT stage. In the Candidate SQL Selection stage, we compute a confidence score for every generated SQL statement across all models and select the one with the highest confidence score as the final output.

Formally, let  $D_i$ denote the structured schema information produced by the Schema Linking procedure in Section 4.2, which achieves a trade-off between recall and precision. This high-quality schema information generated from Schema Linking serves as input to the SQL generation module and significantly enhances SQL accuracy.  Let $ Q^*$ represent a set of diverse SLMs (e.g., Qwen3-4b, DeepSeek-7b, CodeLlama-7b), each adapted via parameter-efficient fine-tuning. The hyperparameter $ k^*,\theta^* $ of Schema Linking and the model parameters are optimized based on the loss function.
\begin{equation}
\min_{ k^*,\theta^* ,\sigma} \frac{1}{|T|} \sum_{i=1}^{|T|} \mathcal{L}_{Q^*} \big( \sigma(q_i, D^*_i), s_i \big)
\end{equation}
where $q_i$ is the natural language question, $D_i^*$ is the Schema Linking output, and $s_i$ is the ground-truth SQL query.

The Candidate SQL Generation stage employs Beam Search algorithm across all supervised fine-tuning SLMs to produce a pool of candidate SQL sequences:
\begin{equation}
H_{sql}
=\bigcup_{m=1}^M\arg\max_{s \in \mathrm{BeamSearch}_k\!\left(Q_m^{*},\,\sigma(q, D^{*})\right)}
P\!\left(s \mid Q_m^{*}, \sigma(q, D^{*})\right)
\end{equation}
$m$ represents different SLMs and $H_{sql}$ indicates the SQL queries in this set are generated by different SLMs. We utilize Beam Search rather than greedy search to decode SQL statements, as the former avoids premature convergence to local optima. Experiments indicate that leveraging Beam Search significantly enhances the quality of SQL generation for individual model by capturing sequences with higher global probabilities.

Finally, in the Candidate SQL Selection stage, we evaluate each candidate SQL statement $h_{sql(i)}\in H_{sql}$ using its reasoning trace confidence which defined as the negative average log-probability of its tokens:
\begin{equation}
TC(h_{sql(i)}) = \sqrt{\frac{1}{|h_{sql(i)}|} \sum_{j=1}^{|h_{sql(i)}|} (T_{sql(j)}-T_{sql\:avg(i)})^2}
\end{equation}
$T_{sql(j)}$ represents the token confidence of the j-th token in the SQL query.  we select the lowest-confidence (greatest model certainty) SQL query as the final output:
\begin{equation}
h_{sql}^*=\arg\min_{h_{sql(i)}\in H_{sql}}TC(h_{sql(i)})
\end{equation}
The trace confidence scores from different SLMs use different scales. To fix this, we use the "z-score" method to normalize the trace confidence scores so they can be compared fairly.  This integrated framework ensures that both Schema Linking and SQL generation benefit from ensemble diversity, uncertainty-aware decoding, and confidence-guided selection, leading to robust and accurate Text-to-SQL parsing.

\section{Experiments}

In this section, we first present the experimental settings. We conduct comparative experiments using three SLMs with fewer than 1 billion parameters.  We evaluated CHS-SQL against diverse Text-to-SQL strategies, including other PEFT method,  few-shot learning prompting,  CoT method, and multi-agent collaboration frameworks. The results demonstrate that our proposed method achieves superior performance when deployed on LLM with fewer than one billion parameters. Furthermore, we investigated the effectiveness of various schema linking approaches and their downstream impact on SQL generation. Our findings indicate that the schema linking approach in CHS-SQL yields more effective results, primarily because it achieves an optimized precision-recall trade-off. Subsequent ablation studies further confirm that this improvement in schema linking is directly correlated with the enhanced accuracy of the final SQL statements.

\subsection{Setting}
In this section, we first introduce the experimental settings. We conduct comparative experiments using three mainstream large language models (LLMs) with fewer than 1 billion parameters. First, we compare CHS-SQL against existing Schema Linking methods to demonstrate that our approach achieves superior performance in terms of the precision–recall trade-off. Next, we evaluate CHS-SQL against state-of-the-art parameter-efficient fine-tuning methods based on LLM architectures on the Text-to-SQL task. Furthermore, we perform ablation studies to validate the effectiveness of the key components of our method.

\subsubsection{Dataset}
We employ two widely used benchmark datasets for Text-to-SQL evaluation. The first is Spider, introduced by Yale University in 2018, which spans multiple databases and domains. Compared to simpler datasets such as WikiSQL, Spider features more complex SQL constructs and advanced operations—including  GROUP BY, ORDER BY, HAVING, and multi-table JOIN—making it significantly more realistic and challenging. The Spider dataset comprises 8,659 training examples, 1,034 development examples, 2,147 test examples, and schema definitions for 372 distinct databases.
The second dataset is BIRD (BIg Bench for Large-scale Database Grounded Text-to-SQL Evaluation), a pioneering cross-domain benchmark that explicitly investigates the impact of extensive real-world database contents on Text-to-SQL parsing. BIRD contains over 12,751 unique question–SQL pairs and 95 large-scale databases with a total size of 33.4 GB.

\subsubsection{Models}
We select three sub-billion-parameter LLMs for our experiments: CodeLlama-7B, Qwen-7B, and DeepSeek-Coder-6.7b-Instruct. Quantized versions of these models can all be trained and inferred on a single NVIDIA RTX 4090 GPU with 24 GB of memory. To ensure a fair comparison across methods and minimize the confounding effect of fine-tuning, we uniformly apply LoRA across all models and algorithms. Specifically, we set the LoRA target modules to \verb|q_proj| and \verb|v_proj| for all models. For each model, all algorithms share identical hyperparameter settings. For instance, when using CodeLlama-7b, we fix the batch size to 2, learning rate to $5e^{-5},$ LoRA rank (r) to 64, and LoRA alpha to 32.

\subsubsection{Metrics}
To evaluate the quality of the Schema Linking process, we adopt the \textbf{F1 }score, which aims to maximize the recall of GT (ground-truth) tables and GT columns while minimizing redundant schema elements.
For the SQL generation stage, we report two standard evaluation metrics:
\begin{itemize}
    \item \textbf{EM (Exact Match)}: measures whether the predicted SQL query exactly matches the ground-truth SQL in all components;
    \item \textbf{EX (Execution Accuracy)}: evaluates whether the predicted SQL and ground-truth SQL queries produce identical execution results on the underlying database.
    \item \textbf{VES (Valid Efficiency Score)}: measures the execution efficiency of the predicted SQL query compared to the ground-truth SQL, rewarding queries that are not only correct but also computationally efficient on the database.
\end{itemize}

\subsection{Text2SQL Performance with SLMs}

\begin{table*}[t]
\centering
\caption{Text-to-SQL Performance}
\label{tab:schema_linking_results}
\small
\begin{adjustbox}{max width=\textwidth}
\begin{tabular}{p{6.3cm}lcc}
\toprule
\textbf{Method} & \textbf{Model} & \multicolumn{2}{c}{\textbf{SPIDER DEV}} \\
\cmidrule(lr){3-4}
 &  & \textbf{EX} & \textbf{EM} \\
\midrule

DTS-SQL & CodeLlama-7B & 0.715 & 0.700 \\
DTS-SQL & DeepSeek-Coder-6.7b-Instruct & 0.753 & 0.706 \\
DTS-SQL & Qwen3-4B & 0.674 & 0.651 \\
MAC-SQL & Qwen3-4B & 0.472 & 0.120 \\
Dail-SQL & Qwen3-4B & 0.519 & 0.484 \\
DIN-SQL & Qwen3-4B & 0.450 & 0.392 \\
CHS-SQL & Qwen3-4B & 0.766 & 0.738 \\
CHS-SQL & CodeLlama-7B & 0.736 & 0.697 \\
CHS-SQL & DeepSeek-Coder-6.7b-Instruct & 0.769 & 0.733 \\

CHS-SQL & \makecell[l]{DeepSeek-Coder-6.7b-Instruct \\ \& Qwen3-4b} & \textbf{0.773} & \textbf{0.739} \\
\midrule
\textbf{Method} & \textbf{Model} & \multicolumn{2}{c}{\textbf{SPIDER TEST}} \\
\cmidrule(lr){3-4}
 &  & \textbf{EX} & \textbf{EM} \\
\midrule
DTS-SQL & CodeLlama-7B & 0.707 & 0.672 \\
CHS-SQL & CodeLlama-7B & 0.741 & 0.684 \\
\midrule
\textbf{Method} & \textbf{Model} & \multicolumn{2}{c}{\textbf{BIRD DEV}} \\
\cmidrule(lr){3-4}
 &  & \textbf{EX} & \textbf{VES} \\
\midrule
DTS-SQL & CodeLlama-7B-int4 & 0.654 & 0.595 \\
CHS-SQL & CodeLlama-7B-int4 & 0.742 & 0.664 \\
\bottomrule
\end{tabular}
\end{adjustbox}
\end{table*}

The CHS-SQL method uses Beam Search during the schema linking stage. This helps the model find more relevant information (recall) and prevents it from getting stuck on the first, most obvious answer (local optima). By adding a way to check the model’s  internal confidence, we can filter out low-quality or unnecessary data in both the schema linking and SQL generation stages. This significantly improves the final results.

We conduct a comparative study between CHS-SQL and the baseline method DTS-SQL on the Text-to-SQL task, with a focus on evaluating the effectiveness of our method when applied to fine-tuning  SLMs.  The experimental results not only demonstrate that our novel Text-to-SQL approach achieves superior performance, but also confirm that Schema Linking outputs that strike a well-balanced between precision and recall significantly enhance the accuracy of downstream SQL generation.

According to the results in Table 1, we tested CHS-SQL on the Spider Dev dataset using three different small models (under 1 billion parameters). Compared to our baseline (DTS-SQL), CHS-SQL improved the EX score by over 2\% across all models, and the EM score by nearly 2\% on some. When using the Qwen3-4B model, CHS-SQL performed even better, beating the baseline by more than 6\% in both EX and EM scores.

As shown in Table 1, we also compared our approach with other methods like MAC-SQL, Dail-SQL, and DIN-SQL, which rely on prompt engineering or multi-agent systems. These methods performed much worse, with scores often 20\% lower than ours. We believe there are two main reasons for this:
\begin{enumerate}
    \item \textbf{Instruction following}: SLMs often struggle to follow complex prompts. For example, in DIN-SQL, the model sometimes fails to output just the SQL code. Instead, it adds extra analysis or puts the code inside specific tags (like \verb|'''SQL'''|), which requires extra work to fix. Currently, there is not much research on prompt engineering specifically for these small language models.
    \item \textbf{Reasoning power}: SLMs naturally have weaker reasoning skills and less general knowledge than very large models, which makes Text-to-SQL tasks harder for them. Our experiments show that fine-tuning (PEFT) is a much more effective way to improve their performance on specific tasks.

\end{enumerate}

The experiments in Table 1 prove that CHS-SQL is more accurate than other methods when using a single SLM. Furthermore, the data shows that when multiple SLMs work together, the results improve even more. On the Spider Dev dataset, our scores reached 75.4\% (EX) and 75.2\% (EM). This is an impressive result, as it is nearly as good as models with hundreds of billions of parameters.

Finally, CHS-SQL also performed very well on other datasets, proving it works in different scenarios. On the Spider Test set, our EX score improved by 3.4\% over the baseline. On the BIRD dataset, the advantage was even clearer, with both EX and EM scores increasing by more than 6.9\%.

\subsection{Precision-recall Trade-off Schema Linking Performance}

In this subsection, we employ the Precision Rate, Recall Rate and F1 Score to evaluate and compare the performance of CHS-SQL against various baseline methods in the Schema Linking stage. In Table 2, "T-P", "T-R", and "T-F1" represent the Precision, Recall, and F1 Score for ground truth tables, respectively. Similarly, "C-P", "C-R", and "C-F1" denote the Precision, Recall, and F1 Score for ground truth columns.

Specifically, we compare CHS-SQL with the baseline method DTS-SQL across SLMs. In contrast to DTS-SQL, which only evaluates the retrieval of  GT(ground-truth) tables, our evaluation additionally assesses the ability of each method to retrieve GT columns, providing a more comprehensive analysis of schema coverage. Furthermore, to enable a broader and more rigorous evaluation, we also include comparisons with RESD-SQL(a BERT-based pre-trained model)  on its Schema Linking results. While our method uses a generative model to score and filter sequences, RESD-SQL uses a discriminative model to directly score how relevant each table and column is to the question. This multi-faceted comparison allows for a thorough evaluation of the effectiveness of CHS-SQL.

\begin{table}[H]

\caption{The impact of Schema Linking on generating SQL}
\label{tab:schema_linking_metrics_fit}
\scriptsize
\setlength{\tabcolsep}{3pt}
\renewcommand{\arraystretch}{1.15}

\resizebox{\linewidth}{!}{%
\begin{tabular}{|p{5.8cm}|l|c|c|c|c|c|c|c|c|}
\hline
\textbf{Method} & \textbf{Model} &
\textbf{T-P} & \textbf{T-R} & \textbf{T-F1} &
\textbf{C-P} & \textbf{C-R} & \textbf{C-F1} &
\makecell{\textbf{Spider Dev} \\ \textbf{EX}} & \makecell{\textbf{Spider Dev} \\ \textbf{EM}}  \\
\hline
Fine-tuning SLM & CodeLlama-7B & 0.944 & 0.964 & 0.954 & 0.873 & 0.901 & 0.887 & 0.696 & 0.673 \\

 &  &  &  &  &  &  &  &  & \\

DTS-SQL  & CodeLlama-7B & 0.944 & 0.964 & 0.954 & 0.263 & 0.970 & 0.415 & 0.731 & 0.720 \\

 &  &  &  &  &  &  &  &  & \\

\makecell[l]{CHS-SQL \\ (with filtered tables and filtered columns)} & CodeLlama-7B & 0.661 & 0.991 & 0.793 & 0.459 & 0.975 & 0.624 & 0.695 & 0.643 \\

 &  &  &  &  &  &  &  &  & \\

CHS-SQL & CodeLlama-7B & 0.661 & 0.991 & 0.793 & 0.192 & 0.992 & 0.322 & \textbf{0.736} & \textbf{0.697} \\

 &  &  &  &  &  &  &  &  & \\

 RESD-SQL Schema Linking & CodeLlama-7B & 0.432 & 1.0 & 0.604 & 0.155 & 0.998 & 0.268 & 0.691 & 0.671 \\

 &  &  &  &  &  &  &  &  & \\

All Tables chosen and All Columns chosen  & CodeLlama-7B & 0.334 & 1.0 & 0.501 & 0.112 & 1.0 & 0.201 & 0.660 & 0.630 \\

 &  &  &  &  &  &  &  &  & \\

\makecell[l]{CHS-SQL \\ (with TEXT/NUM/DATE match)} & CodeLlama-7B & 0.495 & 1.0 & 0.662 & 0.155 & 1.0 & 0.269 & 0.722 & 0.712 \\

 &  &  &  &  &  &  &  &  & \\
 
DTS-SQL  & DeepSeek-Coder-6.7b & 0.937 & 0.964 & 0.950 & 0.260 & 0.968 & 0.410 & 0.713 & 0.708  \\

 &  &  &  &  &  &  &  &  & \\
 
CHS-SQL & DeepSeek-Coder-6.7b & 0.795 & 0.991 &0.882& 0.232 & 0.992 & 0.376 & \textbf{0.729} & \textbf{0.721}\\

\midrule
\textbf{Method} & \textbf{Model} &
\textbf{T-P} & \textbf{T-R} & \textbf{T-F1} &
\textbf{C-P} & \textbf{C-R} & \textbf{C-F1} &
\makecell{\textbf{Spider Test} \\ \textbf{EX}} & \makecell{\textbf{Spider Test} \\ \textbf{EM}}  \\
\midrule
DTS-SQL & CodeLlama-7B & 0.941 & 0.935 & 0.938 & 0.288 & 0.940 & 0.441 & 0.715 & 0.678 
\\
 &  &  &  &  &  &  &  &  & \\

CHS-SQL & CodeLlama-7B & 0.618 & 0.993 &0.762& 0.198 & 0.993 & 0.330 & \textbf{0.741} & \textbf{0.684}\\
\hline
\end{tabular}%
}

\end{table}

To compare how different Schema Linking methods affect the final SQL generation results, they all used the same Beam Search method to pick the most likely SQL query in all experiments. As shown in Table 2, the "Fine-tuning SLM" method uses a fine-tuned SLM to identify the necessary database schema subset. It uses a simple greedy search without any special filtering. While this method has the lowest recall finding fewer relevant schema subset, it has very high precision and a high F1 score. Its final EX score is close to 70\%, which proves that a high F1 score helps improve SQL generation.

The DTS-SQL strategy also relies on a fine-tuned SLM. It identifies the tables first and then includes all columns from those tables as input for the SQL model. Compared to the first method, this approach has the same accuracy for tables but a 7\% higher recall for columns. However, its precision dropped by 60\%. Even so, it improved the final EX score by 3.5\% and the EM score by 4.7\%. Since we used Beam Search for SQL generation here just like the other methods, this version of DTS-SQL performs better than the standard version shown in Table 1.

The "CHS-SQL(with filtered tables and filtered columns)" strategy uses both Beam Search and Trace Confidence filtration to filter out unnecessary tables and columns. Compared to the standard CHS-SQL (which only filters tables), this method has higher precision and a better F1 score. It also has better recall for both tables and columns than the first two methods. However, we noticed its EX and EM scores were actually lower. There are two reasons: first, it failed to find a good balance (trade-off) between precision and recall; second, although it found many correct schema subset, they were scattered across different samples, meaning fewer total samples were completely correct.

Standard CHS-SQL has the highest recall for both tables and columns compared to the other three strategies, but it has the lowest precision and F1 score. Despite this, it achieved the highest EX and EM scores. This is because the combination of Beam Search and Trace Confidence filtration helps CHS-SQL find the perfect precision-recall balance, which is key to better SQL generation.

Next, we looked at the RESD-SQL Schema Linking strategy. Compared to CHS-SQL, it has higher recall but lower precision. As a result, its final SQL scores were even worse than the first method (the simple model with no special strategy).

We also tested a "All Tables chosen and All Columns chosen" approach, where we gave the model all tables and columns in the database without filtering. As expected, this led to the worst results because there was too much useless information.

The "CHS-SQL (with TEXT/NUM/DATE match)" strategy adds a text-matching step. It matches words, numbers, and dates from the user’s question with the database content. This achieved 100\% recall for both tables and columns. Its EX and EM scores both exceeded 70\%.

In conclusion, Our data shows that the key to better SQL generation is to first ensure high recall for tables and columns, and then improve precision. Finding the right precision-recall trade-off is the most important factor. In our tests, methods with very high precision (like "Fine-tuning SLM") or nearly 100\% recall (like "RESD-SQL Schema Linking") both performed worse than the balanced CHS-SQL method. Our additional experiments using the DeepSeek-7B model and the Spider Test set also support this conclusion.

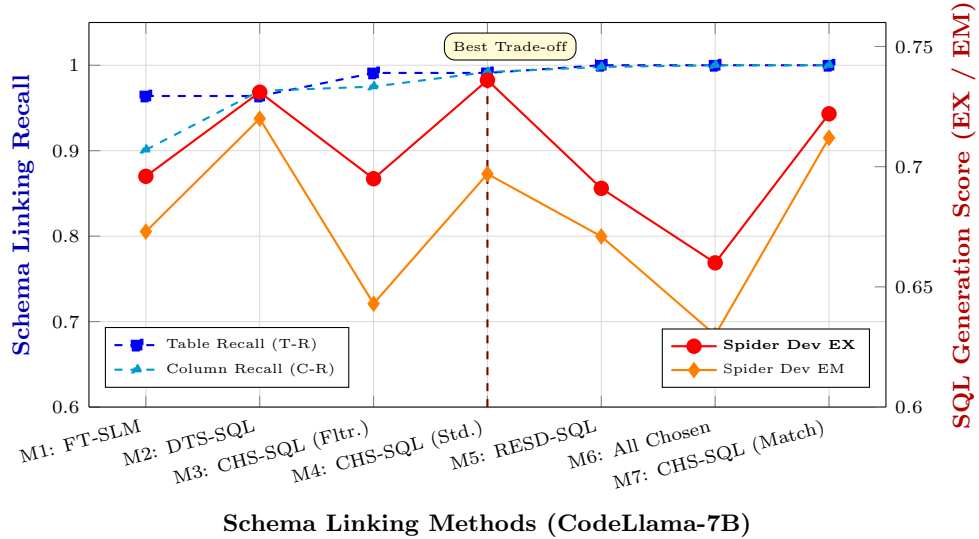
\begin{figure}[htbp]
\centering
\caption{The relationship between Schema Linking performance (Table/Column Recall) and the final SQL generation accuracy (EX/EM) on Spider Dev dataset.}
\label{fig:schema_linking_vs_sql}
\begin{tikzpicture}
\begin{axis}[
    width=1.0\textwidth,
    height=0.55\textwidth,
    xmin=0.5, xmax=7.5,
    ymin=0.6, ymax=1.05,
    axis y line*=left,
    xlabel={\small\textbf{Schema Linking Methods (CodeLlama-7B)}},
    ylabel={\small\textbf{Schema Linking Recall}},
    ylabel style={font=\color{blue!70!black}},
    ticklabel style={font=\scriptsize},
    xtick={1,2,3,4,5,6,7},
    xticklabels={
        {M1: FT-SLM},
        {M2: DTS-SQL},
        {M3: CHS-SQL (Fltr.)},
        {M4: CHS-SQL (Std.)},
        {M5: RESD-SQL},
        {M6: All Chosen},
        {M7: CHS-SQL (Match)}
    },
    x tick label style={rotate=15, anchor=north east},
    grid=both,
    grid style={line width=.1pt, draw=gray!10},
    major grid style={line width=.2pt, draw=gray!30},
    legend style={at={(0.02,0.05)}, anchor=south west, font=\tiny, cells={anchor=west}}
]

\addplot[color=blue, mark=square*, thick, dashed] coordinates {
    (1, 0.964) (2, 0.964) (3, 0.991) (4, 0.991) (5, 1.000) (6, 1.000) (7, 1.000)
};
\addlegendentry{Table Recall (T-R)}

\addplot[color=cyan!80!black, mark=triangle*, thick, dashed] coordinates {
    (1, 0.901) (2, 0.970) (3, 0.975) (4, 0.992) (5, 0.998) (6, 1.000) (7, 1.000)
};
\addlegendentry{Column Recall (C-R)}
\end{axis}

\begin{axis}[
    width=1.0\textwidth,
    height=0.55\textwidth,
    xmin=0.5, xmax=7.5,
    ymin=0.60, ymax=0.76,
    axis y line*=right,
    axis x line=none,
    ylabel={\small\textbf{SQL Generation Score (EX / EM)}},
    ylabel style={font=\color{red!70!black}},
    ticklabel style={font=\scriptsize},
    legend style={at={(0.98,0.05)}, anchor=south east, font=\tiny, cells={anchor=west}}
]

\addplot[color=red, mark=*, thick, mark size=2.5pt] coordinates {
    (1, 0.696) (2, 0.731) (3, 0.695) (4, 0.736) (5, 0.691) (6, 0.660) (7, 0.722)
};
\addlegendentry{\textbf{Spider Dev EX}}

\addplot[color=orange, mark=diamond*, thick, mark size=2.5pt] coordinates {
    (1, 0.673) (2, 0.720) (3, 0.643) (4, 0.697) (5, 0.671) (6, 0.630) (7, 0.712)
};
\addlegendentry{Spider Dev EM}

\draw[red!50!black, thick, dashed] (axis cs:4,0.60) -- (axis cs:4,0.736);
\node[draw, rectangle, fill=yellow!20, font=\tiny, rounded corners] at (axis cs:4.2,0.75) {Best Trade-off};

\end{axis}
\end{tikzpicture}
\end{figure}

The experimental results presented in Figure 2 clearly demonstrate a critical correlation between the precision-recall balance during the Schema Linking phase and the final downstream Text-to-SQL execution (EX) and exact match (EM) accuracies. The key insights from these observations can be synthesized into the following arguments:  
\begin{itemize}
\item High Recall as a Prerequisite for Downstream Accuracy: A comparison across various configurations reveals that achieving a high recall rate for both tables (T-R) and columns (C-R) serves as a foundational prerequisite for competitive SQL generation performance. For instance, while the "Fine-tuning SLM" method yields exceptionally high precision, its relatively low recall restricts the overall SQL generation capacity, bounding its EX score near 69.6\%. Conversely, strategies that prioritize recall consistently unlock higher performance upper-bounds.  

\item    The Precision-Recall Trade-off Bottleneck: Although high recall is necessary, an unconstrained increase in recall at the expense of precision introduces substantial schema noise, which severely degrades model performance. This is exemplified by the "All Tables chosen and All Columns chosen" baseline; despite achieving a perfect 100\% recall, the absence of filtration exposes the SLM to excessive irrelevant schema elements, culminating in the lowest EX score (66.0\%). Similarly, the "RESD-SQL Schema Linking" method exhibits high recall but suboptimal precision, which ultimately causes its final SQL generation metrics to underperform even the simple baseline.  

\item     Superiority of Balanced Optimization via CHS-SQL: The optimal synergy is achieved when a method successfully navigates the precision-recall trade-off. By leveraging Beam Search to ensure broad schema coverage (high recall) alongside Trace Confidence filtration to prune irrelevant entities, CHS-SQL establishes a well-calibrated candidate schema subset.  
\item     Detrimental Effects of Over-Filtration: Interestingly, when aggressive filtering is applied to both tables and columns—as seen in "CHS-SQL (with filtered tables and filtered columns)"—the schema linking precision and F1-scores increase, yet the final EX and EM scores paradoxically decline. This drop occurs because over-filtering disrupts the delicate trade-off, causing correct schema subset to become scattered across fragmented samples. Consequently, fewer total samples are rendered fully correct during generation, reinforcing that a balanced subset outperforms a hyper-filtered one.  

 \item Maximizing Performance with Hard Feature Enhancements: When text-matching constraints (e.g., q-table-match, q-column-match, and num-date-cell-match) are integrated into CHS-SQL, the framework guarantees a flawless 100\% schema recall while maintaining controlled precision. This robust integration pushes both EX and EM scores beyond the 70\% threshold, validating that a structured approach to stabilizing recall without flooding the context with noise is the definitive pathway to maximizing SLM effectiveness in Text-to-SQL tasks.

\end{itemize}

\subsection{Ablation Experiments}

To validate the effectiveness of CHS-SQL, we conduct a series of ablation studies in Spider Dev dataset with Codellama-7b. Table 3 shows how the CHS-SQL method performs when different components are removed, focusing on the EX and EM scores. CHS-SQL (with three SLMs voting) represents the full process. It uses three models (Qwen3-4b, Codellama-7b and DeepSeek-coder-6.7b) to work together and choose the best SQL query. As expected, this version achieves the highest scores in both EX and EM. CHS-SQL (only with Codellama-7b) uses just one model. This means it skips the step where multiple models vote and filter results using Trace Confidence. Compared to standard CHS-SQL (the voting version), the single model version CHS-SQL that its EX score drops by 1.8\% and the EM score drops by 5.5\%. The following experiments also use the single model version CHS-SQL to test the affect on SQL generation when specific components are removed:
\begin{itemize}
\item   w/o Beam Search in Schema Linking: In this test, we removed Beam Search during the schema linking stage and relied only on the fine-tuned SLM to filter schema subset. This caused the EX score to drop by 4\% and the EM score by 2.4\%.
\item   w/o Trace Confidence Filtration: Here, we stopped using Trace Confidence to filter out redundant table information. This led to a 1\% drop in EX and a 0.4\% drop in EM.
\item   w/o All Columns Chosen: This version uses the Beam Search and Trace Confidence to filter redundant schema subset, but gives all columns directly to the SQL model. In contrast, the standard CHS-SQL first filters extra tables and then selects all columns of filtered tables as the inputs of the SQL generation model. This change caused EX to drop by 4.1\% and EM by 5.4\%.
\item   w/o Schema Linking: This means no filtering was done at all. Every table and column in the database was sent to the SQL generation model. This removed component causes the biggest negative impact, with EX falling by 7.6\% and EM by 6.7\%.
\item   w/o Beam Search in SQL Generation: In this test, we used a simple Greedy Search instead of Beam Search to generate the SQL statements. The EX score decreased by 2.2\% and the EM score by 1.8\%.
\end{itemize}

Together, these ablation experiments provide comprehensive evidence for the individual and synergistic efficacy of each component in CHS-SQL. It confirm that every part of CHS-SQL plays an important role in making SQL generation more accurate.

\begin{table*}[t]
\centering
\caption{Ablation Experiments}
\label{tab:schema_linking_results}
\footnotesize
\resizebox{0.7\textwidth}{!}{
\begin{tabular}{p{5.3cm}cc}
\toprule
\textbf{Method} & \textbf{EX} & \textbf{EM} \\
\midrule
CHS-SQL (with multiple SLMs voting)& 0.773 & 0.739 \\
CHS-SQL (only with Codellama-7b) & 0.736 & 0.697 \\
\multicolumn{1}{>{\raggedleft\arraybackslash}p{5.3cm}}{w/o Beam Search in Schema Linking} & 0.696 & 0.673\\
\multicolumn{1}{>{\raggedleft\arraybackslash}p{4.7cm}}{w/o Trace Confidence Filtration} & 0.726 & 0.693\\
\multicolumn{1}{>{\raggedleft\arraybackslash}p{3.8cm}}{w/o All Columns Chosen} & 0.695& 0.643\\
\multicolumn{1}{>{\raggedleft\arraybackslash}p{3.2cm}}{w/o Schema Linking} & 0.660 & 0.630\\
\multicolumn{1}{>{\raggedleft\arraybackslash}p{5.4cm}}{w/o Beam Search in SQL Generation} & 0.714& 0.679   \\
\bottomrule
\end{tabular}
}
\end{table*}

\section{Discuss}
The experimental results show that filtering out redundant information during Schema Linking improves the performance of SLMs in Text-to-SQL tasks. This matches the findings of previous research. Data from Table 2 reveals that high recall is key to improving SQL accuracy. When the F1 score is the same, a higher recall rate brings more benefits than higher precision. Some Schema Linking strategies in Table 2 even reached 100\% recall, but they did not get the best EX and EM scores because their precision was too low. The data also shows that if recall stays at 100\%, increasing precision will further improve the final SQL accuracy.

In summary, the key to success for SLMs in Text-to-SQL tasks is finding the best precision-recall trade-off. This means filtering out as much redundant information as possible while still finding all the correct schema subset. Our CHS-SQL method introduces an innovative way to balance precision and recall for both tables and columns, which is why it achieves the best performance.

\section{Conclusion}
Our proposed two-stage PEFT framework, integrated with an innovative token generation strategy, achieves strong performance on theText-to-SQL task when applied to SLMs. Although numerous prior works have demonstrated that filtering out redundant schema subset during Schema Linking benefits downstream SQL generation, current mainstream LLM-based approaches primarily rely on the inherent capabilities of the base model and prompt engineering to maximize recall in Schema Linking, without explicitly exploring how to simultaneously improve precision while maintaining high recall.

In this work, we evaluate Schema Linking quality and further observe that high recall in retrieving GT tables and GT columns is essential for ensuring downstream SQL generation quality. Building upon a high-recall candidate set, we then apply a ingenious filtering mechanism to remove redundant tables and columns, thereby enhancing the precision of the final Schema Linking output. This refined approach enables further improvements in SQL generation accuracy.

Our experiments confirm that CHS-SQL effectively achieves a well-balanced between precision and recall in the Schema Linking phase. Moreover, the high-quality Schema Linking results produced by CHS-SQL significantly boost the accuracy of SQL queries generated by SLMs, underscoring the critical importance of high quality schema retrieval inText-to-SQL systems.

\section{Limitation}
In Table 2, some Schema Linking methods already achieve 100\% recall. Although these methods have the lowest precision among all groups, their final results are still close to CHS-SQL. These methods also show a clear trend: when recall for GT tables and GT columns stays at 100\%, higher precision leads to better performance in the final SQL task. This suggests that if we can increase precision while maintaining full recall, we might find an even better precision-recall trade-off than our current CHS-SQL method.Currently, we have not yet found a way to achieve higher precision while guaranteeing 100\% recall. This will be a key area for our future research.

\section*{Acknowledgements} 
This work was supported by the Hainan Provincial Key Research and Development Program of 2025 under Grant No. ZDYF2025GXJS179 (Project Title: \textit{Research, Development, and Application of Artificial Intelligence Technology for Digital Supply Chain in Hainan Free Trade Port}).

\bibliographystyle{plain}
\bibliography{references}

\end{document}